\documentclass{article}
\pdfoutput=1
\usepackage[square,sort,comma,numbers]{natbib}

\usepackage[preprint]{neurips_2020}

\usepackage[pdftex]{graphicx}
\usepackage[utf8]{inputenc} 
\usepackage[T1]{fontenc}    
\usepackage{hyperref}       
\usepackage{url}            
\usepackage{booktabs}       
\usepackage{amsfonts}       
\usepackage{nicefrac}       
\usepackage{microtype}      

\usepackage{color}
\usepackage{mathtools}
\usepackage{algorithm}
\usepackage{algpseudocode}
\newcommand\mdoubleplus{\ensuremath{\mathbin{+\mkern-10mu+}}}

\title{Continual Learning with Deep Artificial Neurons}

\author{%
  Blake Camp\thanks{first author} \\
  Department of Computer Science\\
  Georgia State University\\
  Atlanta, GA 30302 \\
  \texttt{bcamp2@student.gsu.edu} \\
   \And
   Jaya Krishna Mandivarapu \\
   Department of Computer Science\\
   Georgia State University\\
   Atlanta, GA 30302 \\
   \texttt{jmandivarapu1@student.gsu.edu} \\
   \AND
   Rolando Estrada \\
   Department of Computer Science\\
   Georgia State University\\
   Atlanta, GA 30302 \\
   \texttt{restrada1@gsu.edu} \\
}

\begin{document}

\maketitle

\begin{abstract}
   Neurons in real brains are enormously complex computational units.  Among other things, they're responsible for transforming inbound electro-chemical vectors into outbound action potentials, updating the strengths of intermediate synapses, regulating their own internal states, and modulating the behavior of other nearby neurons. One could argue that these cells are the only things exhibiting any semblance of real intelligence. It is odd, therefore, that the machine learning community has, for so long, relied upon the assumption that this complexity can be reduced to a simple sum and fire operation. We ask, might there be some benefit to substantially increasing the computational power of individual neurons in artificial systems?  To answer this question, we introduce Deep Artificial Neurons (DANs), which are themselves realized as deep neural networks. Conceptually, we embed DANs inside each node of a traditional neural network, and we connect these neurons at multiple synaptic sites, thereby vectorizing the connections between pairs of cells.  We demonstrate that it is possible to meta-learn a single parameter vector, which we dub a neuronal phenotype, shared by all DANs in the network, which facilitates a meta-objective during deployment. Here, we isolate continual learning as our meta-objective, and we show that a suitable neuronal phenotype can endow a single network with an innate ability to update its synapses with minimal forgetting, using standard backpropagation, without experience replay, nor separate wake/sleep phases.  We demonstrate this ability on sequential non-linear regression tasks.
   
\end{abstract}

\section{Introduction}

Humans, and indeed most organisms, can be regarded as programmable machines.  These machines come prepacked with with certain remarkable, innate abilities.  One such ability is the capacity to continually acquire new knowledge without forgetting the past.  This is commonly referred to as Continual Learning, or Learning without Forgetting \cite{ANML, EWC, WGD, meta_representations_for_CL, learning_without_forgetting}.  Unfortunately, this trait has largely eluded artificial learning systems; in particular,  Deep Artificial Neural Networks (ANNs).  While memory in real brains is not fully understood, it is generally accepted that memories and knowledge are, at least temporarily, encoded in the connections (synapses) between neurons \cite{principles_of_neural_science}.  ANNs have demonstrated incredible abilities across a variety of domains by making use of some clever algorithms which excel at finding good configurations of synaptic weights for particular tasks \cite{origin_of_deep_learning}.  Historically, however, they have been susceptible to a phenomenon known as catastrophic forgetting, which is caused by weight updates that fail to preserve prior knowledge \cite{catastrophic_forgetting}.  This problem has persisted because the most proven algorithm by which to train deep networks, Gradient Descent by Backpropagation, has been notoriously incapable of computing memory-preserving synaptic updates without access to historical data.  We suspect that the persistence of this problem may have resulted from our collective infatuation with finding good synaptic weights for \textit{networks}, in lieu of discovering specific responsibilities of \textit{neurons}.

As noted in \cite{Spike_based_causal_inference_for_weight_alignment}, "any learning system that makes small changes to its parameters will only improve if the changes are correlated to the gradient of the loss function."  This wonderful insight lays bare the necessity that alternatives to Gradient Descent by Backpropagation must, at a minimum, correlate to the primary objective which it explicitly computes.  This has led many to consider whether or not the brain is performing something similar to Backpropagation, despite some lingering questions regarding it's biological plausibility \cite{Spike_based_causal_inference_for_weight_alignment, random_feedback_weights}. Given that this debate persists, and given the fact that Backpropagation works so well, it stands to reason that there may exist a way to leverage it's strengths, while minimizing it's weaknesses. 

It has been argued "that much of an animal’s behavioral repertoire is not the result of clever learning algorithms...but arises instead from behavior programs already present at birth...(and that) these programs arise through evolution, are encoded in the genome, and emerge as a consequence of wiring up the brain" \cite{critique_of_pure_learning}.  However, this intuition, apt as it may be, might raise more questions than it answers. Which priors, for example, need to be encoded in the genome? How are those priors realized in the organism?  Since most contemporary AI research is concerned with how to best go about finding good synaptic weights and architectures for specific tasks, it makes sense that this bias would be reflected in recent literature \cite{MAML, ANML, meta_representations_for_CL, WGD}.  On the other hand, very little work has been done to investigate the potential of learning powerful priors for neurons.

Real neurons are really complex \cite{dynamical_systems, dendritic_morphogenesis, segregated_dendrites, single_neurons_solve_mnist, single_cortical_neurons_as_DNNs, principles_of_neural_science, on_intelligence}. They transform electro-chemical vectors into outbound action potentials and synaptic updates.  The strengths of their connections are defined not by single scalar values, but by high-dimensional vectors representing the state of chemical concentrations, electrical potentials, and proximity. They grow new appendages (dendrites), and some cells even physically migrate in the brain \cite{neuronal_migration}. Conversely, artificial neurons, with few exceptions, are extraordinarily simple models which reduce this enormous biological complexity to an elementary sum and fire operation.  We ask, might there be some benefit to dramatically increasing the computational power of individual neurons in artificial neural networks?  Is it possible that the cause of a lot of innate behavior has been hiding in plain sight; not in the connectome, but inside the neurons themselves?

To answer this question, we introduce Deep Artificial Neurons, or DANs. Conceptually, we embed DANs inside each node of a traditional neural network, and we connect these neurons at multiple synaptic sites, thereby vectorizing the connections between pairs of cells.  We demonstrate that it is possible to meta-learn a single parameter vector, which we dub a neuronal phenotype, shared by all DANs in the network, which facilitates a meta-objective during deployment. Here, we isolate continual learning as our meta-objective, and we show that a suitable neuronal phenotype can endow a single network with an innate ability to update its synapses with minimal forgetting, using standard backpropagation, without experience replay, nor separate wake/sleep phases.  In doing so, we demonstrate the benefits of embedding innate structure inside the \textit{neurons} of a larger plastic network. Our proposal offers the benefits of Backpropagation, specifically a guarantee of behavioral improvement, resulting from correlation with the gradient of the loss function, while also minimizing its susceptibility to catastrophic forgetting. We demonstrate this capacity on sequential non-linear regression tasks.

\begin{figure}
  \centering
  \includegraphics[width=0.8\linewidth]{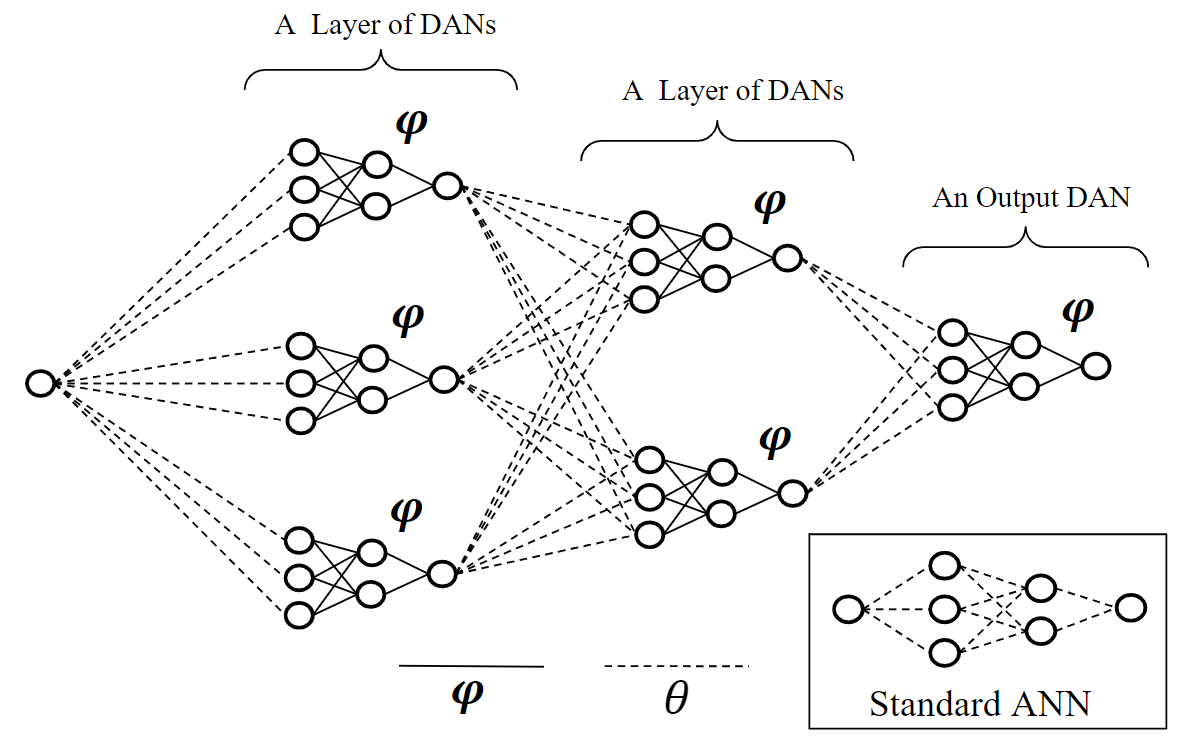}
  
  \caption{A Network of Deep Artificial Neurons (DANs).  DANs are connected to one another by parameters $\theta$, which can be regarded as vectorized synapses, or VECs. All DANs share parameters $\varphi$, which we dub a neuronal phenotype.  The strength of the connection existing between any 2 connected neurons is therefore a function of the state of the n-dimensional synaptic vector. }
  \label{fig:diagram}
\end{figure}

\section{Related Work}

There is at least some consensus that memories and knowledge are stored in the connections (synapses) between neurons \cite{principles_of_neural_science, dynamical_systems}.  The malleability of these connections is known as synaptic plasticity, and the rules governing how the strengths are updated is known as learning.  ANNs have excelled largely because Backpropagation is good at learning configurations of synaptic weights for immediate, well-defined tasks \cite{origin_of_deep_learning}. However, the algorithm has historically failed in non-iid settings where the training distribution evolves over time \cite{EWC}.  

On the other hand, the plasticity dynamics of real neurons, even on short time frames, are much more complex than many in the machine learning community like to acknowledge. Even more biologically faithful attempts to explain the underlying update rules have failed to adequately model the dynamics required for long-term knowledge preservation \cite{burst_dependent_synaptic_plasticity}. 

Let us review some material in 3 areas crucial to our interpretation of the problem:  Complex Neurons and Synapses, Innate Structural Priors and Meta-Learning, and Continual Learning and Memory.

\subsection{Complex Neurons and Synapses}

There have been some very recent attempts to model more complex neurons and synapses.  However, suffice it to say, it has still not been fully explored in the artificial intelligence literature. Some have sought to manually enforce dendritic segregation as a means to facilitate multi-plexing of feed-forward and feed-back signals in deep artificial networks \cite{segregated_dendrites, numenta_synapses}.  Deep Complex Networks aimed to increase the complexity of the signals being propagated by introducing complex-valued connections and activations \cite{deep_complex_networks}. In \cite{fast_weights}, the authors also investigated the potential benefits of 2 sets of synaptic weights; one for fast retention of knowledge, and another for memory over longer time horizons.  Beniaguev et al. showed that a multi-layer ANN can mimic much of the functionality of real neurons \cite{single_cortical_neurons_as_DNNs}.  Aguera y Arcas et al. advocated for learning complex priors for neurons and synapses \cite{blaise_aguera}, and a similar approach was advocated in \cite{online_neural_updates_learning_to_remember} specifically for continual learning. Jones et al. demonstrated that single neurons may be computationally capable of solving sophisticated vision tasks \cite{single_neurons_solve_mnist}. Mordvintsev et al. showed that it may be beneficial to think of individual cells as networks with self-organizing properties and high-dimensional message passing capabilities \cite{NCA}.  In \cite{mplp}, the authors expanded on this notion by formalizing a Message Passing Learning Protocol, whereby learning is facilitated through the communication of messages, realized as vectors, passed amongst nodes.    

\subsection{Innate Structural Priors and Meta-Learning}
Many in the machine learning community have begun to at least acknowledge that efficient learning in real brains may result, not from clever learning algorithms, but from innate priors which have arisen through evolution, and which are generally kept fixed during intra-life deployment \cite{critique_of_pure_learning}.  Weight-Agnostic Neural Networks demonstrated that there exist architectures, not specific weight configurations, which exhibit out-of-the-box propensity for certain tasks \cite{weight_agnostic_NN}.  The Lottery Ticket Hypothesis has shown that there often exist smaller, easily optimizable networks embedded in the weight matrices of much larger networks \cite{lottery_ticket}, again hinting at the existence of optimal structure for specific tasks. As such, here have been numerous attempts to meta-learn a good prior over the weight distributions and architectures of ANNs.  MAML aims to find parameters suitable for fast adaptation and few-shot learning \cite{MAML}. In \cite{one_policy}, the authors propose a technique to meta-learn a single, shared policy network which is distributed throughout the anatomy of a robot, allowing one network to control the behavior of multiple body-parts via message-passing.

\subsection{Continual Learning}
Generally speaking, machine learning optimization algorithms rely upon the \textit{iid} assumption, which asserts that the training and test distributions need to be (approximately) the same. For this reason, the most reliable technique to date for overcoming catastrophic forgetting is one which explicitly enforces this assumption, often referred to as experience replay (ER).  ER aims to ensure that the system is exposed to data drawn uniformly from the underlying task distribution/s. Continual Learning, however, requires memory retention even in settings where the current training distribution may not be representative of the total, historical data distribution; thus requiring a \textit{non-iid} assumption. Broadly speaking, 4 categories of techniques have emerged to combat catastrophic forgetting: (1) regularization-based approaches \cite{EWC, orthogonalGradientDescent, regularizationShortcomings}, (2) knowledge-compression, or capacity expansion \cite{meta_consolidation_for_cl, selfnet, Dynamically_Expandable_Networks, CL_with_hypernetworks}, (3) experience replay and/or external memory slots \cite{memory_networks, atari_with_deep_rl, experience_replay_for_cl, neural_turing_machines}, and (4) meta-learned representations and update rules \cite{ANML, meta_representations_for_CL, feedback_and_local_plasticity, WGD, online_neural_updates_learning_to_remember}.   

\begin{figure}
  \centering
  \includegraphics[width=0.8\linewidth]{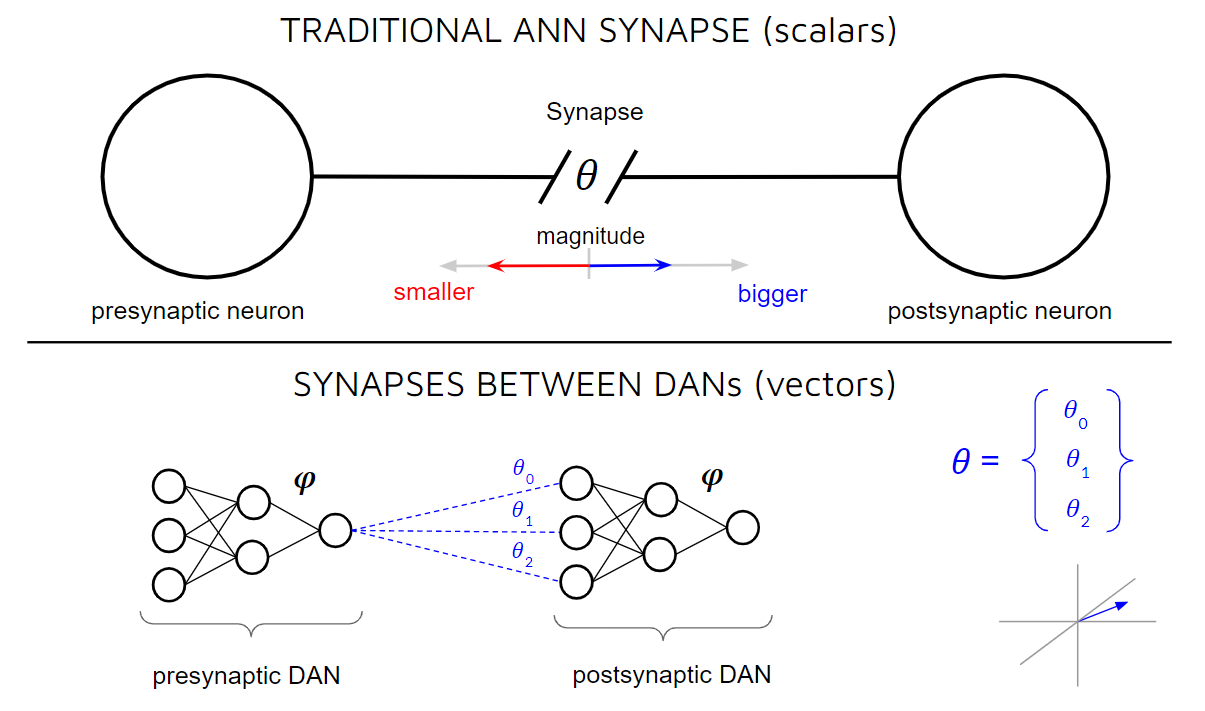}
  
  \caption{Synapses in Traditional ANNs are defined by single, scalar values.  Synapses in a Network of DANs are vectors.  These Vectorized Synapses, or VECs, may represent multiple connections between pairs of neurons at different synaptic sites, or multi-variable neurotransmitter concentrations and electrical potentials at a single synapse.  The strength of the connection between 2 neurons is therefore an explicit (non-linear) function of the magnitude and orientation of this synaptic vector.}
  \label{fig:synapses}
\end{figure}

There has been a lot of recent interest in meta-learning a good solution to the problem of continual learning \cite{WGD, ANML, meta_representations_for_CL, feedback_and_local_plasticity, online_neural_updates_learning_to_remember}.  This is in stark contrast to more traditional approaches which have attempted to hand-craft algorithmic solutions, e.g. \cite{EWC, Dynamically_Expandable_Networks, orthogonalGradientDescent}.  We draw strongest inspiration from a promising approach known as Warped Gradient Descent (WGD).  WGD mitigates catastrophic forgetting by learning warp parameters $\omega$, realized as warp-\textit{layers}, which are interleaved between the standard layers of a neural network. These warp parameters are held \textit{fixed} during deployment, allowing the rest of the network to learn, without forgetting, using standard backpropagation.  As the name implies, parameters $\omega$ warp activations in the forward pass, and the gradients in the backwards pass. The technique is promising because it offers guarantees of convergence, and a path towards continual-learning at scale. In contrast to WGD, we show in our initial experiments that it is possible to learn a single, small network $\varphi$, constituting a single neuronal phenotype, that can be distributed throughout a larger plastic network, and which also mitigates catastrophic forgetting.






\begin{algorithm}
    \caption{Meta-Learning a Neuronal Phenotype for Continual Learning}\label{alg:meta_training_alg}
    \begin{algorithmic}[1]
        \Require{$\mathit{p}(\mathcal{T})$: distribution over target functions} 
        \Require{$\alpha, \gamma$: learning rate hyperparameters}  
        \Require{inner\_steps: number of inner loop steps}
            \State $\theta \gets \theta_0, \varphi \gets \varphi_0$: randomly initialize the model 
            \While{not done}  
                \State Sample a Target Function $\mathcal{T} \sim \mathit{p}(\mathcal{T})$
                \For{sub-task $st$ in $\mathcal{T}$}
                    \For{$t$ in inner\_steps}
                        \State Perform an update: $\theta_{t+1}\varphi_{t+1} \xleftarrow{}\theta_{t} \varphi_{t}- \alpha \nabla_{\theta_{t}} \gamma \nabla_{\varphi_{t}}\mathcal{L}_{\mathit{st}}^{t}$ \Comment{Equation \ref{eq:VECs_update}}
                    \EndFor
                    \State $\mathcal{H}_{t+1} \gets$ data from sub\_tasks[0,...,$st$] $\subset \mathcal{T}$ 
                    \State{Compute Memory-Loss over $\mathcal{H}_{t+1}$} 
                    \State{Update Phenotype: 
                     $\varphi_{t+1}^{\prime} = \varphi_{t+1} - \gamma\nabla_{\varphi_{t+1}}\mathcal{L}_{\mathcal{M}}^{{\mathcal{H}_{t+1}}}$  \Comment{Equation \ref{eq:optimal_meta_update}}}
                \EndFor
                \State $\theta \gets \theta_0$: reset VECs to initialization 
            \EndWhile
       
    \end{algorithmic}
\end{algorithm}

\section{Model}
\label{sec:model}
Deep Artificial Neurons, or DANs, are themselves realized as multi-layer neural networks. For the purposes of demonstration, consider a DAN instantiated as a 2-layer neural network, with a single layer of hidden nodes, and a single output node (i.e. the output activation of the neuron).  In practice, we apply a \textit{tanh} non-linear activation to the output of the hidden layer, as well as to the output layer.  See the bottom of Figure \ref{fig:synapses} for an illustration.  Let n\_channels denote both (1) the size of the input vector to this network, and (2) as we will see, the number of connections between pairs of DANs.  Conceptually, we can distribute this single DAN amongst all nodes of a traditional neural network.  Figure \ref{fig:diagram} offers an illustration of how to convert a standard ANN into a Network of DANs with n\_channels=3. 

More generally, consider the topology of a standard, fully-connected, feed-forward neural network with $l$ layers of nodes, and let $n_{l}$ denote the number of nodes in layer $l$.  Let $l_0$ be a special case, denoting the layer of input nodes, which are \textit{not} DANs. We can convert this topology to a Network of DANs in the following way.  For each layer of nodes, up to but not including the layer of output nodes, we instantiate a layer of Vectorized Synapses, or VECs, as a standard, fully-connected weight-matrix $\theta_{l}$ with dimensions $n_{l} \times (n_{l+1} \times n\_channels)$.  VECs connect layers of DANs to one another.  Feed-forward propagation of a signal along these connections is therefore facilitated in the standard way, by computing the dot product of the activation vector $\sigma_{l\_out}$ from the previous layer and this layer of VECs $\theta_{l}$.  This yields a large input vector $\sigma_{(l+1)\_in}$, to be processed by the DANs in the next layer:

\begin{equation}
\label{eq:FF_input_vector}
    \sigma_{(l+1)\_in} = \sigma_{l\_out} \cdot \theta_{l} = \sum_i^{n_{l+1}}{\theta_{i}^{j}\sigma_{l\_out}^j + b_i},
\end{equation}

 where $j$ denotes the index of nodes in layer $l$, and $i$ is the index of nodes in layer $l+1$. Note that in practice we do \textit{not} apply a non-linear activation function to this vector.  Rather, slices of the raw dot product $\sigma_{(l+1)\_in}$ are fed as input to the DANs in the next layer. That is, since all DANs share parameters $\varphi$, the same DAN model processes each slice $s_{n}^{l+1}$ of $\sigma_{(l+1)\_{in}}$, which denotes the number of DANs $n$ in layer $l+1$.  Said another way, the input vector to the layer of DANs is sliced into $n_{l+1}$ equally sized sub-vectors. The output vector of a layer of DANs is obtained by passing each of these separate slices through the DAN, and concatenating the resulting activations. Let $\mdoubleplus$ denote the concatenation operation:

\begin{equation}
\label{eq:FF_output_vector}
    \sigma_{(l+1)\_out} = \sigma_{(l+1)\_out}\mdoubleplus  \varphi(s_{n}^{l+1}) \textrm{\space\space} \forall \textrm{\space\space} \textrm{slices in layer} \textrm{\space\space} \mathit{l}+1
\end{equation}

\begin{figure}
  \centering
  \includegraphics[width=0.8\linewidth]{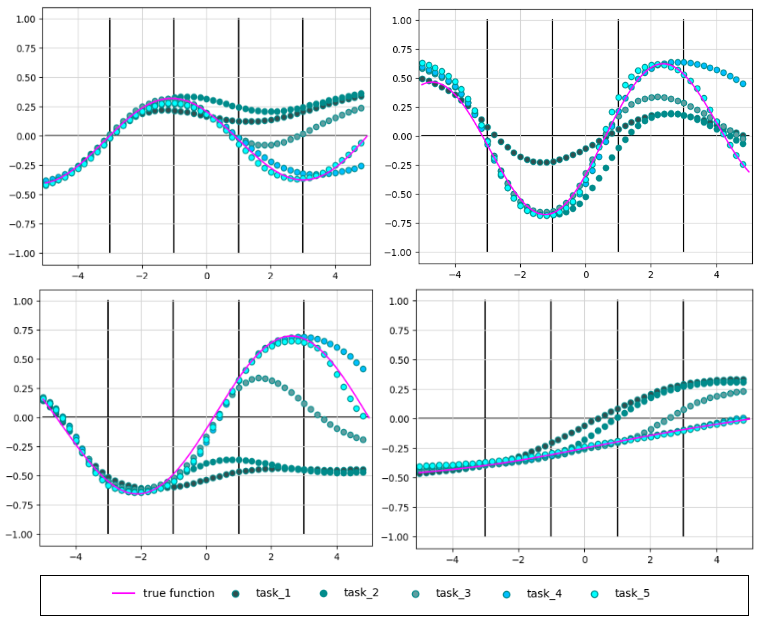}
  
  \caption{Continual Learning during Deployment of 4 non-linear functions, each divided into 5 sub-tasks, using a meta-learned neuronal phenotype which is held fixed. VECs are updated with standard backpropagation. Each uniquely colored scatter plot depicts the predictions of the model over the whole function after performing 100 updates on data from the current sub-task only.  In other words, the darkest plot represents the model's predictions over the whole function after training only on task\_1 $[-5, -3)$ .  During the next stage of learning, the model performs 100 updates on data from task\_2 only $[-3,-1)$.  The lightest plot (cyan) depicts the model's predictions after the last round of learning: 100 updates on data from task\_5 $[3,5]$.  As we can see, the model retains a good fit over the whole function even when it learns these sub-tasks in a sequential manner.}
  \label{fig:CL_examples}
\end{figure}

In practice, we also use skip connections, inspired by Deep Residual Networks \cite{deep_residual_learning} and \cite{WGD}, in order to facilitate efficient learning.  Skip connections are realized as additional layers of VECs $\theta_{skip(j,k)}$, with dimensions $n_{j} \times $$(n_{k} \times n\_channels)$,  which bypass layers of DANs by providing a direct pathway from nodes in layer $j$ to layer $k$, where $k = j+2$.  This allows some information to be sent directly downstream, without being subjected to processing by the intermediate layer of DANs.  When using skip conenctions, the input vector to a layer of DANs in layer $k$ is obtained by summing the vector computed in Equation  \ref{eq:FF_input_vector} with the vector signal traveling along $\theta_{skip(j,k)}$:

\begin{equation}
\label{eq:Skip_input_vector}
    \sigma_{k\_in}^{\prime} = \sigma_{k\_in} + (\sigma_{j\_out} \cdot \theta_{skip(j,k)})
\end{equation}

The result is a complete model, comprised of 2 distinct sets of parameters: VECs, parameterized by $\theta$, and DANs, parameterized by $\varphi$.  As we will show in coming sections, VECs are intended to be fully plastic at all times.  The parameters of our DANs, our so-called neuronal \textit{phenotype}, are meta-learned and then held fixed during deployment.  

As in \cite{WGD}, we leverage the benefits of a unique set of parameters $\varphi$ which can be shown to warp the gradients applied to another set of parameters $\theta$ in order to prevent catastrophic forgetting.  In contrast to WGD, however, we show that a single, small network, parameterized by $\varphi$, is sufficient to facilitate our meta-objective, rather than unique, separate layers of warp parameters. Additionally, we feel that our approach offers an additional layer of biological plausibility, and might help to explain some of the behavior and responsibilities of real neurons.

\section{Problem Formulation}
\label{sec: problem_formulation}
In this work, we set out to investigate whether it is possible to meta-learn a single parameter vector $\varphi$, shared by all DANs in the model, which can mitigate catastrophic forgetting in a network which learns using standard backpropagation during deployment. We call this parameter vector a neuronal phenotype, since it defines the behavior of each neuron in the network, and is kept fixed during intra-lifetime deployment.  To this end, we formulated an experiment similar to the one proposed in \cite{WGD}, and originally in \cite{MAML}.  More specifically, we consider the problem of sequential non-linear regression, wherein a model must try to fit to a complete function, when exposed to data from only part of that function in distinct time-intervals.  In other words, it must learn the complete function in a \textit{piece-wise}, or incremental manner, since it cannot revisit data to which it was exposed during previous intervals.  As in \cite{WGD}, we split the input domain $[-5, 5]$ $\subset$ $\mathbb{R}$ into 5 consecutive sub-intervals, which correspond to 5 distinct tasks.  Task\_1 therefore corresponds to the sub-function falling within $[-5,-3)$; Task\_2 corresponds to the sub-function within $[-3,-1)$, and so on.  The model is exposed to Tasks 1 through 5 in sequential manner.  During each sub-task, the network is exposed to 100 data points, drawn uniformly from the current task window.  That is, during Task\_1 the model performs 100 updates on data sampled from $[-5,-3)$.  In our experiments, we perform 1 update on every sample, equating to a batch size of 1. Sub-tasks are thus defined by their respective windows in the input domain.

We slightly modify the target functions used in \cite{WGD}. We define a task sequence by a target function that is a mixture of two sine functions with varying amplitudes, phases, and x-offsets.  At the beginning of each meta-epoch, we randomly sample two amplitudes $\alpha_{(0,1)} \in$  (0,2), phases $\rho_{(0,1)} \in$ (0, $\pi / 3)$, and x-offsets $\phi_{(0,1)} \in [-5,5]$.  Summing two such sine functions yields a target function of the form:

\begin{equation}
\label{eq:target_functions}
    y = \alpha_0 sin((\rho_0 x) + \phi_0) + \alpha_1 sin((\rho_1 x) + \phi_1)  
\end{equation}

Afterwards, we discard any target functions from the training distribution meeting the following criteria: $y_{max} > .8$; $y_{min} > -.8$; $y_{max} - y_{min} < .4$.  This yields a final input domain of $[-5,5]$ and output range of $[-.8, .8]$.

\section{Methodology}
\label{sec:methodology}

Meta-Learning, generally speaking, is typically concerned with optimizing some meta-objective over a distribution of tasks in order to attain some innate proficiency at comparable tasks likely to be encountered during a separate, deployment phase. Therefore, most meta-learning algorithms employ an inner-loop/out-loop framework, wherein optimization over several specific tasks occurs in the inner-loop, and proficiency at the meta-objective is evaluated and optimized in the outer-loop. Inspired by Warped Gradient Descent (WGD) \cite{WGD}, we adopt a such an approach, and wish to meta-learn parameters which facilitate continual learning during deployment.

During meta-training, we randomly sample target functions of the form defined in Equation \ref{eq:target_functions}.  These target functions are split into 5 sub-tasks, as explained in Section \ref{sec: problem_formulation}.  We deploy our model on each target function, and sub-tasks are encountered sequentially. Optimization over a single sub-task is done by performing backpropagation on both sets of parameters, $\theta$ and $\varphi$, using the Loss over the current sub-task.  This is known as an inner-loop epoch.  At the end of each inner-loop epoch, we quantify the Meta-Loss over subtasks [0,...,\textit{cur}], where \textit{cur} is the current sub-task.  Meta-Optimization is done by performing backpropagation on parameters $\varphi$ \textbf{\textit{only}}, using the Meta-Loss.  Repetition of this process over a sequence of 5 sub-tasks, given the current target function, is known as a meta, or outer-loop epoch.  At the end of each outer-loop epoch, we sample a new target function, and repeat the process.

More formally, let the Historical Learning Trajectory $\mathcal{H}_t$ represent the dataset [$x_0$, $x_1$, …, $x_t$] comprised of all data encountered by the system, prior to and including timestep $t$.  We have VECs, parameterized by $\theta$, and DANs, parameterized by $\varphi$, which together define the complete Model.  

Note that since DANs are distributed throughout the network, the gradients for VECs $\nabla_\theta$ depend on parameters $\varphi$, and that the meta-gradient $\nabla_\varphi$ depends on parameters $\theta$.  This is true, since each set of parameters $\theta$ and $\varphi$ are factors of both gradients. 

We can define a Model State at timestep $t$ as $\theta_{t}\varphi_{t}$.  Given a new sample at timestep $t+1$, this Model State will result in a measurable Task Loss $\mathcal{L}_{\mathcal{T}}$, for which we can compute a gradient:

\begin{equation}
\label{eq:task_loss_gradient}
 \nabla_{\theta_{t}\varphi_{t}} \mathcal{L}_{\mathcal{T}}^{t+1}
\end{equation}

Note that we can factor this gradient into its distinct components:

\begin{equation}
\label{eq:factored_task_loss_gradient}
 \nabla_{\theta_{t}\varphi_{t}} \mathcal{L}_{\mathcal{T}}^{t+1} =  \nabla_{\theta_{t}}\nabla_{\varphi_{t}}\mathcal{L}_{\mathcal{T}}^{t+1}
\end{equation}

This is desirable since we may want to assign separate learning rates to each set of parameters.  For instance, let $\alpha$ denote the learning rate for parameters $\theta$, and let $\gamma$ denote the learning rate for parameters $\varphi$. Since we update both $\theta$ and $\varphi$ in the inner loop, when learning individual sub-tasks, performing an inner-loop update on data from the current task at timestep $t+1$, like so:

\begin{equation}
\label{eq:VECs_update}
 \theta_{t+1}\varphi_{t+1} \xleftarrow{} \theta_{t}\varphi_{t} - \alpha \nabla_{\theta_{t}}\gamma\nabla_{\varphi_{t}}\mathcal{L}_{\mathcal{T}}^{t+1}
 \end{equation}

...results in the new Model State $\theta_{t+1}\varphi_{t+1}$.  Note that this update, $\theta_{t+1}\varphi_{t+1} \xleftarrow{} \theta_{t}\varphi_{t}$, may have caused forgetting over $\mathcal{H}_{t+1}$, which now includes the latest data sample $x_{t+1}$.  

We can quantify the Memory Loss over $\mathcal{H}_{t+1}$, defined as $\mathcal{L}_{\mathcal{M}}^{{\mathcal{H}_{t+1}}}$. This constitutes the meta-loss, which we wish to minimize in order to \textit{facilitate} our meta-objective during inner-loop deployment.  

Specifically, we seek an optimal neuronal phenotype, defined by a single parameter vector $\varphi^*$, shared by all DANs, which would have resulted in the least amount of forgetting over $\mathcal{H}_{t+1}$.  Said another way, had the original state of the model been $\theta_{t}\varphi_{t}^*$, instead of  $\theta_{t}\varphi_{t}$, then the inner loop update would have been:

\begin{equation}
\label{eq:optimal_theta_update}
 \theta_{t+1}^*\varphi_{t+1}^* \xleftarrow{} \theta_{t}\varphi_{t}^* - \alpha \nabla_{\theta_{t}}\gamma\nabla_{\varphi_{t}^*}\mathcal{L}_{\mathcal{T}}^{t+1}
\end{equation}

This would have resulted in an alternative Model State $\theta_{t+1}^*\varphi_{t+1}^*$, ideally resulting in less forgetting than that originally induced by $\theta_{t}\varphi_{t}$.

Therefore, we calculate the Memory Loss across $\mathcal{H}_{t+1}$, using the current model state $\theta_{t+1}\varphi_{t+1}$, and compute the gradient w.r.t. this quantity:

\begin{equation}
\label{eq:meta_gradient}
 \nabla_{\theta_{t+1}\varphi_{t+1}}\mathcal{L}_{\mathcal{M}}^{{\mathcal{H}_{t+1}}}
\end{equation}

By taking a step towards $\varphi_{t}^{*}$, we update the phenotype $\varphi$, and in the process attempt to minimize the meta-loss.  We can do this by factoring the gradient and isolating the update to $\varphi$ only:

\begin{equation}
\label{eq:optimal_meta_update}
 \varphi_{t+1}^{\prime} = \varphi_{t+1} - \gamma\nabla_{\varphi_{t+1}}\mathcal{L}_{\mathcal{M}}^{{\mathcal{H}_{t+1}}} \textrm{\space\space\space  s.t.  \space\space\space} \varphi_{t+1}^{\prime} \approx \varphi_{t}^*
\end{equation}

The full meta-training procedure is outlined in Algorithm \ref{alg:meta_training_alg}. 

After meta-training is completed, the model is \textit{deployed}. DAN parameters $\varphi$ are held \textbf{\textit{fixed}}, and the model is obligated to learn continually, without forgetting, using standard backpropagation.  That is, $\theta$ update normally, while DANs remain fixed.  We offer empirical validation of our approach in the next section.

\section{Experiments and Results}
\label{sec:experiments}

\begin{figure}
  \centering
  \includegraphics[width=0.8\linewidth]{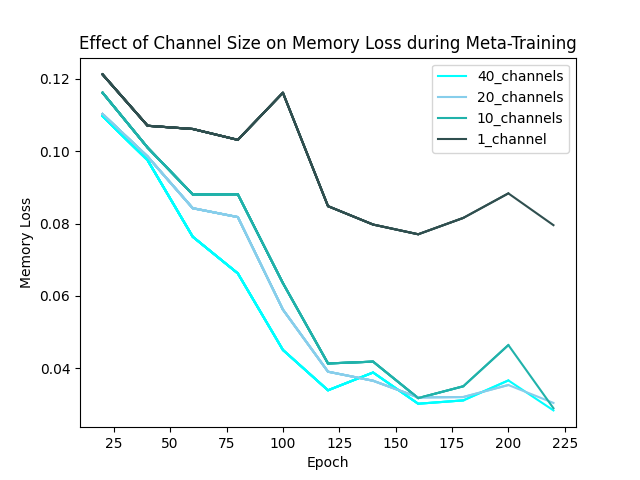}
  
  \caption{Vectorizing the connections between pairs of neurons has a clear impact on  the efficiency with which the network learns to minimize the Memory Loss during Meta-Training. The number of channels denotes both the size of the input vector fed to each DAN, and the number of connections that each presynaptic neuron has to each postsynaptic neuron. }\label{fig:channels}
\end{figure}

To validate our approach, we performed experiments on tasks defined in Section \ref{sec: problem_formulation}.  For all experiments, we used a network topology of 1 input node, 2 hidden layers of 40 nodes each, and a single output node. Recall that, apart from the single node in the input layer, each node represents a DAN, and the topology is therefore \textit{converted} to a Network of DANs. To this topology, we added 2 skip layers, as described in Section \ref{sec:model}: from layer 0 to layer 2, and also from layer 1 to layer 3.  The DAN itself is a 3 layer neural network with $n\_channels$ input nodes, followed by a hidden layer with 15 nodes, another hidden layer with 8 nodes, and a single output node, parameterized by $\varphi$. We applied  \textit{tanh} activations to the hidden and output layers of the DAN.  For all experiments except that depicted in Figure \ref{fig:channels}, we set $n\_channels=40$. For Meta-Training, we set the learning rate for VECs parameters $\theta = .001$, and the learning rate for DAN parameters $\varphi=.0001$.  

Figure \ref{fig:CL_examples} shows the ability of a meta-trained model to learn continually during \textbf{\textit{deployment}}; when it encounters tasks in a sequential manner, and is obligated to retain a good fit over previous sub-tasks, even though it is exposed to data from each task only once. During this experiment, DAN parameters $\varphi$ were held fixed, and the network learns by using standard backpropagation to update VECs $\theta$.

\begin{figure}
  \centering
  \includegraphics[width=0.8\linewidth]{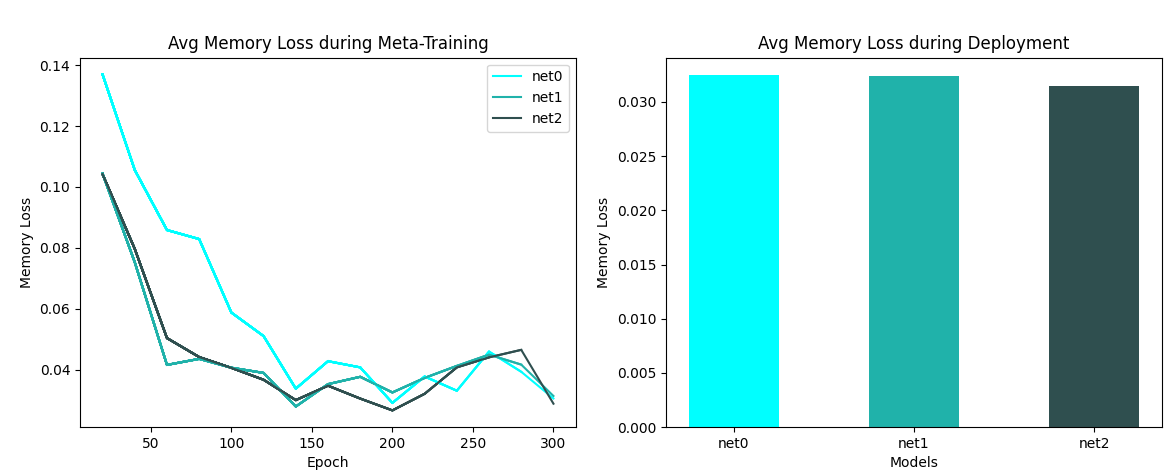}
  
  \caption{Model definitions: \textbf{net0} uses a single, shared phenotype (cell-type); \textbf{net1} uses one phenotype for each layer; \textbf{net2} does not enforce parameter sharing among any DANs. (left) Minimization of the Memory-Loss during Meta-Training.  (right) Avg Memory Loss during Deployment is nearly equal for all models, and nearly identical to the loss achieved near the end of meta-training, $\approx .03$ (mean squared error) across the full task-trajectory after learning 5 sub-tasks in sequence. }
  \label{fig:meta_training_1}
\end{figure}

Figure \ref{fig:meta_training_1} depicts minimization of the Memory-Loss, our meta-objective, during meta-training.  We found that the model converges relatively quickly, requiring only 200-300 meta-epochs to find a suitable phenotype, though this is likely due to task simplicity.  Additionally, as the Figure shows, we sought to isolate the effect of using a single set of parameters for DANs in the whole network.  To do this, we compared 3 models: one which used a single parameter vector for all DANs (net0: a single phenotype throughout the network), another which used a separate parameter vector for each \textit{layer} of DANs (net1: phenotypes unique to each layer), and a third which did not enforce any parameter sharing amongst DANs (net2; 81 unique DANs in the network). 

\begin{figure}
  \centering
  \includegraphics[width=0.8\linewidth]{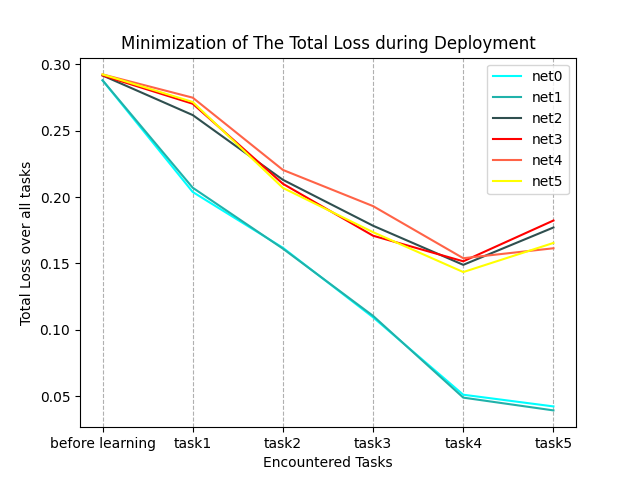}
  
  \caption{Model definitions: \textbf{net0} uses a single, meta-learned phenotype, shared by all DANs, fixed during deployment; \textbf{net1} uses the same meta-learned single phenotype as \textbf{net0}, but it is fully plastic during deployment (updates to the $\varphi$ are allowed); \textbf{net2} uses a random, shared phenotype, fixed during deployment; \textbf{net3} uses a random, shared phenotype, fully plastic during deployment; \textbf{net4} uses random, but completely unique DANs (no parameter sharing), fixed during deployment; \textbf{net5} uses random, but unique DANs, fully plastic during deployment.  Once before learning begins, and after training on each successive task, the Total-Loss over the complete function is calculated. Clearly, the meta-learned phenotype outperforms random DANs.}
  \label{fig:deployment_total_loss}
\end{figure}

\begin{figure}
  \centering
  \includegraphics[width=0.8\linewidth]{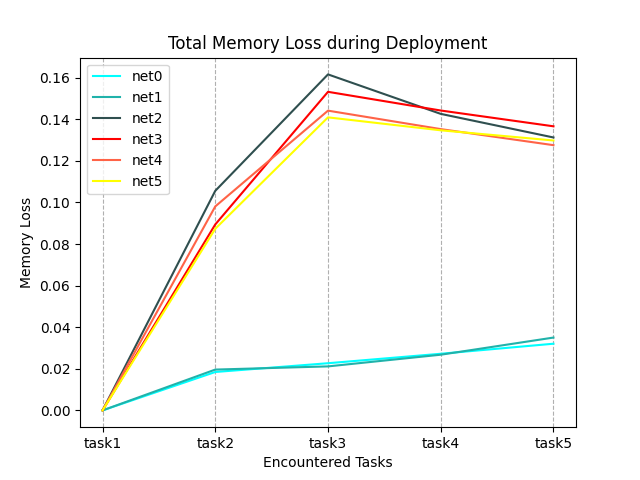}
  
  \caption{Model definitions: same as those in Figure \ref{fig:deployment_total_loss}. Total amount of Memory-Loss experienced by different models during deployment. Clearly, the meta-learned phenotype outperforms random DANs.}
  \label{fig:deployment_memory_loss}
\end{figure}

Figures \ref{fig:deployment_total_loss} and \ref{fig:deployment_memory_loss} compare the abilities of various models during deployment, when they are confronted with tasks in a sequential manner, and obligated to learn continually.  Specifically, we sought to verify whether the meta-learning procedure was indeed endowing the DANs with an innate ability to assist in learning without forgetting.  These plots confirm that hypothesis, showing that a meta-learned phenotype outperforms random parameter vectors, regardless of whether they are fully plastic during deployment, or fixed.  Figure \ref{fig:deployment_total_loss} shows that as each sub-task is learned, the model's prediction error over the whole function is minimized. Figure \ref{fig:deployment_memory_loss} depicts the amount of total memory loss experienced by each model after each stage of learning.  

Finally, we investigated the effect of the size of $n\_channels$ on the ability of the model to minimize Memory-Loss during meta-training.  Specifically, we asked, is there indeed a benefit to vectorizing the connections between pairs of DANs, and in the process increasing the size of the input to each DAN?  Figure \ref{fig:channels} shows that the answer to that question was also yes. The plot shows that as the number of (1) connections between pairs of neurons and (2) the size of the input to each DAN grows, the speed, or efficiency, with which the Memory-Loss is minimized is increased.  In other words, vectorized connections accelerated optimization of our meta-objective.  

\section{Conclusions and Future Work}

Neurons in real brains are enormously complex computational units.  In this work, we offered a framework for thinking about artificial neurons as much more powerful functions, realized as deep artificial networks, which can be embedded inside larger plastic networks. We showed that it is possible to meta-learn a single parameter vector for such a model that, when distributed throughout a larger plastic network and fixed, facilitates a meta-objective during deployment.  In the process, we hope to inspire a deeper understanding about the responsibilities of neurons in both artificial neural networks, as well as real brains.  In future work, we plan to investigate the potential of DANs in real-world vision and reinforcement-learning settings, as well as the possibility of optimizing several meta-objectives at once.

\small



\bibliographystyle{plain}

\bibliography{references}
\end{document}